\documentclass[letterpaper, 10 pt, conference]{ieeeconf}  

\IEEEoverridecommandlockouts                              

\usepackage{graphics} 
\usepackage{epsfig} 
\usepackage{draftwatermark}
\usepackage{amsmath} 
\usepackage{amssymb}  
\usepackage[ruled,linesnumbered]{algorithm2e}
\usepackage{todonotes}

\usepackage{tikz}
\usepackage{textcomp}
\usepackage{hyperref}
\usepackage{lipsum}

\newcommand\copyrighttext{%
  \footnotesize \textcopyright 2012 IEEE. Personal use of this material is permitted.
  Permission from IEEE must be obtained for all other uses, in any current or future
  media, including reprinting/republishing this material for advertising or promotional
  purposes, creating new collective works, for resale or redistribution to servers or
  lists, or reuse of any copyrighted component of this work in other works.}
\newcommand\copyrightnotice{%
\begin{tikzpicture}[remember picture,overlay]
\node[anchor=south,yshift=10pt] at (current page.south) {\fbox{\parbox{\dimexpr\textwidth-\fboxsep-\fboxrule\relax}{\copyrighttext}}};
\end{tikzpicture}%
}

\title{\LARGE \bf
On The Effect of Vibration on Shape Sensing of Continuum Manipulators Using Fiber Bragg Gratings*
}

\author{Shahriar Sefati$^{1, 2, 3}$, {\it Member, IEEE}, Farshid Alambeigi$^{1, 2}$, {\it Member, IEEE}, Iulian Iordachita$^{1,2}$, \\ {\it Senior Member, IEEE}, Russell H. Taylor$^{1,3}$, {\it Life Fellow, IEEE}, and Mehran Armand$^{1,2,4}$, {\it Member, IEEE}
\thanks{*Research supported by NIH/NIBIB grant R01EB016703 and Johns Hopkins internal funds.}
\thanks{$^{1}$S. Sefati, F. Alambeigi, I. Iordachita, R. H. Taylor, and M. Armand are with the Laboratory for Computational Sensing and Robotics, Johns Hopkins University, Baltimore, MD, USA (sefati@jhu.edu, falambe1@jhu.edu, iordachita@jhu.edu, rht@jhu.edu, mehran.armand@jhuapl.edu).
}%
\thanks{$^{2}$S. Sefati, F. Alambeigi, I. Iordachita, and M. Armand are with the Department of Mechanical Engineering, Johns Hopkins University, Baltimore, MD, USA (sefati@jhu.edu, falambe1@jhu.edu, iordachita@jhu.edu, mehran.armand@jhuapl.edu).
}%
\thanks{$^{3}$S. Sefati, and R. H. Taylor are with the Department of Computer Science, Johns Hopkins University, Baltimore, MD, USA (sefati@jhu.edu, rht@jhu.edu).
}%
\thanks{$^{4}$M. Armand is with the Johns Hopkins University Applied Physics Laboratory, Laurel, MD, USA (mehran.armand@jhuapl.edu).
}%
}

\begin{document}

\SetWatermarkText{Accepted for ISMR 2018}
\SetWatermarkScale{0.4}

\maketitle
\copyrightnotice
\thispagestyle{empty}
\pagestyle{empty}

\begin{abstract}

Fiber Bragg Grating (FBG) has shown great potential in shape and force sensing of continuum manipulators (CM) and biopsy needles. In the recent years, many researchers have studied different manufacturing and modeling techniques of FBG-based force and shape sensors for medical applications. These studies mainly focus on obtaining shape and force information in a static (or quasi-static) environment. In this paper, however, we study and evaluate dynamic environments where the FBG data is affected by vibration caused by a harmonic force e.g. a rotational debriding tool harmonically exciting the CM and the FBG-based shape sensor. In such situations, appropriate pre-processing of the FBG signal is necessary in order to infer correct information from the raw signal. We look at an example of such dynamic environments in the less invasive treatment of osteolysis by studying the FBG data both in time- and frequency-domain in presence of vibration due to a debriding tool rotating inside the lumen of the CM.

\end{abstract}


\section{INTRODUCTION}

Due to their compact size, flexibility, and capability to provide real-time feedback, Fiber Bragg Gratings (FBG) have grasped great attention in medical robotic applications, where space for sensor integration is limited and sensor flexibility is required. Examples of such applications include continuum manipulators (CM) \cite{manipulatorReview}, biopsy needles \cite{abayazid2013} and catheters \cite{catheter}. Further, the real-time FBG data stream is an advantage of using FBG-based shape sensors over medical imaging modalities, since the continuous use of these modalities e.g. X-ray is limited due to great radiation exposure for the patient and the surgical team \cite{liu2015large}. In addition, FBG-based shape sensors do not require a direct line of sight, as opposed to tracking systems such as optical trackers, which makes them a good candidate for sensing in CMs and biopsy needles inside the patient's body. It should be noted that different medical applications (e.g. needle insertion \cite{roesthuis2014three}, osteolysis \cite{alambeigi2016design} and osteonecrosis \cite{alambeigiOsteoNec}) require the FBG sensors to function properly in different environments when interacting with soft and hard tissues, which may exert static or dynamic external forces on the sensor.

In the past, many researchers have studied FBG-based shape and force sensing in CMs and biopsy needles. These studies span from complications and their respective solutions in manufacturing and design of the FBG-based sensors \cite{sefati2016fbg,yi2012spatial}, to error and output accuracy evaluation in different applications \cite{OFDR}-\cite{henkenAccuracy}. Several groups have reported work on FBG-based shape sensing of biopsy needles for measuring bending deflections of the needle in soft tissue \cite{misraNeedle}-\cite{iordachitaNeedle}. 
Many groups have also studied the potential use of FBGs in shape, force and torque sensing of CMs \cite{ryu2014fbg}-\cite{misraForce}. They have reported manufacturing challenges and respective solutions in designing such sensors, as well as developing mathematical models for interpreting the FBG data to meaningful shape and force information. 

\begin{figure}
  \centering
    \includegraphics[scale=0.38]{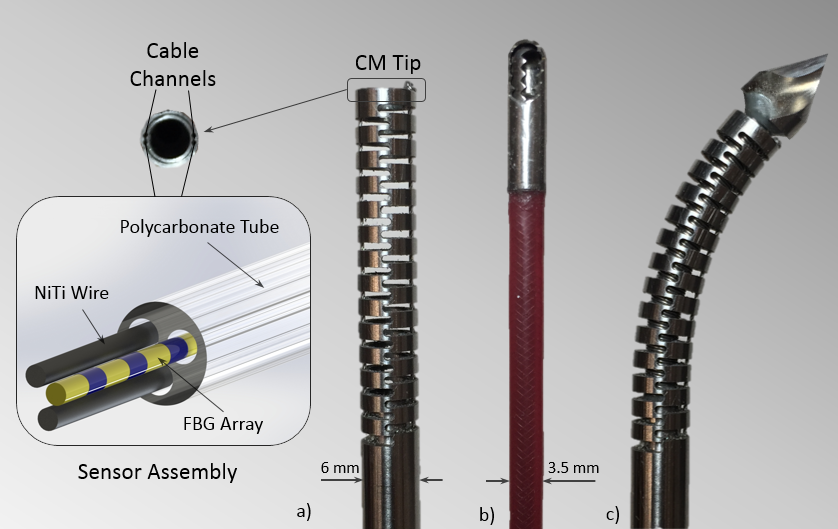}
      \caption{ a) Continuum manipulator tip cross section demonstrating the cable and sensor channels. Schematic of the FBG-based shape sensor consisting of an FBG array and two NiTi wires inserted into the Polycarbonate tube, b) soft lesion debriding tool developed for treatment of osteolysis, and c) hard tissue debriding tool inserted into the CM, developed for osteonecrosis.}
      \label{fig:CDM}
\end{figure}
   
While these studies revealed great potentials of using FBGs in medical applications, to the best of our knowledge, they all focused on static and quasi-static movement of the CM or the biopsy needle (and the FBG sensor, accordingly). As a result, the FBG data only includes changes to the actual shape (or external force) of the CM or biopsy needle, as well as the noise associated with the FBGs and interrogator output. Therefore, a simple low-pass filter might suffice for pre-processing the signal by canceling the noise in the FBG data. However, in presence of vibrational forces exciting the system (FBG sensor combined with the CM or needle), the FBG data will be oscillating depending on the harmonic exciting forces. In these systems, the changes in FBG data is due to not only the typical noise and actual changes in the shape of the CM or biopsy needle, but also the harmonic forces exerted on the system. In addition, other useful information, such as sudden changes in the FBG data due to collision with external objects, might be present in the data. In these situations, blindly using a low-pass filter on the FBG data will cancel the noise as well as other useful information that could have been obtained from the raw data.
   
Examples of medical applications in which the FBG data is exposed to vibration and dynamic movements are the less invasive treatment of osteolysis \cite{alambeigi2018acc}, or osteonecrosis \cite{alambeigiOsteoNec}, in which a CM is used for treatment of soft and hard tissues. For the less invasive treatment of osteolysis and osteonecrosis, we have previously developed a cable-driven CM with an open lumen for passing debriding tools \cite{alambeigi2016design}. In addition, for real time shape sensing of the CM, we have developed and embedded two FBG-based shape sensors into channels on the CM wall  \cite{sefati2016fbg}. As the debriding tools start to rotate inside the lumen of the manipulator, any even small unbalency of the tool causes the system to be excited harmonically depending on the rotational velocity of the tool. 

To fully investigate the effect of vibration on FBG data, in this paper, we carry out several experiments by rotating a debriding tool inside the open lumen of the CM developed for osteolysis, while recording and analyzing the FBG data. More specifically, we investigate the system (CM and the FBG sensor assembly) behavior in different tool rotational velocities $1)$ when no constraint is applied to the system, and $2)$ when the manipulator is bent by actuating the cables. In addition, we study $3)$ the feasibility of using FBGs in presence of vibration and $4)$ how to distinguish the sources of changes in FBG data (whether it is due to actual CM shape changes, or harmonic excitation of the system, or simply noise) by looking at the frequency transform of the time-domain FBG signal. 

The structure of the paper is as follows: section \ref{methods} provides some background on the CM developed for the application of osteolysis, the FBG-based shape sensor, along with the developed debriding tools for treatment of osteolysis. Section \ref{experiments} describes the designed experiments. Section \ref{results} reports and discusses the results from the experiments. Section \ref{conclusion} concludes the paper and mentions the future work.

\begin{figure}
  \centering
    \includegraphics[scale=0.43]{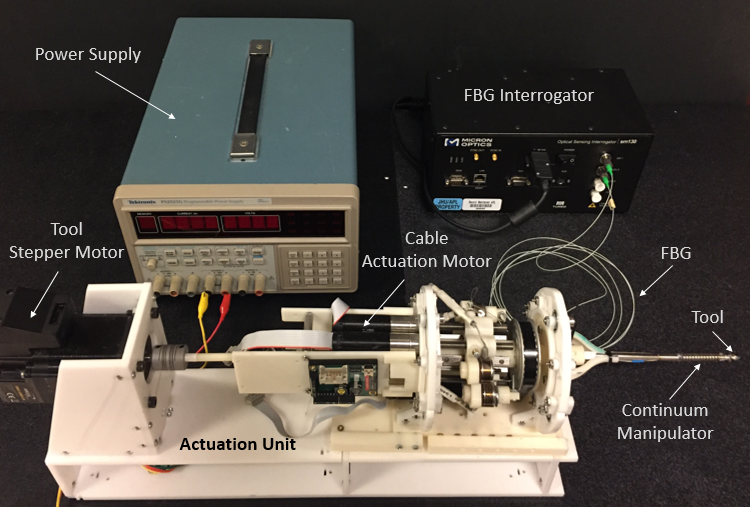}
      \caption{ The experimental setup including an actuation unit for the CM, an FBG interrogator and a power supply. The actuation unit consists of a stepper motor for rotating the tool and DC motors for actuating the CM cables.}
      \label{fig:setup}
\end{figure}

\section{METHODS} \label{methods}

\subsection{Continuum Manipulator Specifications}
Osteolysis (bone degradation) typically occurs due to wear of the polyethylene liner of the acetabular implant after total hip replacement surgery. We have previously developed a CM for the less invasive treatment of osteolysis which is constructed of nitinol (NiTi) with outside and inside diameter of $6$ mm and $4$ mm, respectively (Fig. \ref{fig:CDM}). The CM combined with a robotic arm may be used for other applications such as treatment of articular cartilage injury \cite{forough1}, \cite{forough2} and osteonecrosis of femoral head \cite{alambeigiOsteoNec}. Four small channels (two on each side) with $0.5$ mm diameter are placed on the thin wall of the CM for passing the actuation cables, as well as the FBG-based shape sensors (which we will refer to as the sensor assembly). In the less invasive treatment of osteolysis, the manipulator reaches to the region behind the acetabular component by actuating the cables, while the FBG data is obtained in real-time to provide shape information for feedback control. The flexible debriding tools are then passed through the open lumen to treat the osteolysis.

\subsection{FBG-based Sensor Assembly} \label{Feedback}
The sensor assembly consists of a $0.5$ mm polycarbonate tube with three open lumens, one of which is used for passing and gluing an array of FBG (with three active areas) and the other two are used for passing NiTi wires as substrates (Fig. \ref{fig:CDM}) \cite{sefati2016fbg}. One sensor assembly is embedded in one channel on each side of the CM wall and glued at the distal end (tip) of it. The shape sensor is calibrated so that the wavelength data inferred from the interrogator is translated to curvature at locations with active area. A mathematical model is then used to reconstruct the shape of the CM based on the curvature data \cite{liu2015shape}. As a result, the tip position of the CM can be tracked in real-time and used as a feedback into the controller which actuates the CM cables accordingly.

\subsection{Debriding Tools} \label{tools}
The goal in treatment of osteolysis is to remove the osteolytic lesion formed behind the acetabular implant. The lesion includes both soft \cite{alambeigi2016design} and hard bony parts \cite{alambeigiOsteoNec}, which requires different types of debriding tools to fully clean the region. We have previously developed debriding tools for soft and hard lesion removal which happens in different stages of osteolysis treatment (Fig. \ref{fig:CDM}). 

Each of the developed tools is more optimized in lesion debriding when working in certain ranges of rotational velocities. For soft lesion tools, we have reported performance measurements for rotational velocities up to $200$ rpm, with optimal performance at around $70$ rpm \cite{alambeigi2016design}. Similarly for hard lesion tools, we have observed optimal performance in high rotational velocities close to $2250$ rpm \cite{alambeigiOsteoNec}. It should be noted that these velocities are tool-specific, depending on the geometry and properties (e.g. mass and moment of inertia) of the tool. Therefore, for a new tool, the velocity for optimal debriding should be characterized.

\subsection{Shape Sensing in the Presence of Vibration} \label{motivation}

Previous studies by our group focused on more robust shape sensor manufacturing methods as well as better accuracy in tip estimation in static and quasi-static situations (when no tools were rotating inside the manipulator). However, as soon as a debriding tool starts rotating within the CM, the FBG data begins to show rapid changes depending on the rotational velocity. 
In the presence of vibration, a change in the FBG data might be caused by sources other than the actual changes in the shape of the continuum manipulator. Examples of the sources causing changes in FBG data are: $1)$ actual manipulator shape changes, $2)$ shape sensor vibration, $3)$ external constraints (e.g. contact with obstacles) imposed on the manipulator, and $4)$ noise. Due to small clearance between the shape sensor and the manipulator sensor channel, the vibration might be affecting the shape sensor differently compared to the manipulator itself. Therefore, part of the changes in FBG data might be due to the vibration of the shape sensor (and not the manipulator), which will result in incorrect shape reconstruction of the CM. Moreover, the rapid changes in FBG data, rules out the possibility of inferring other useful information from the data such as collision detection, since decoupling the contribution of each source to changes in FBG data in time domain is not trivial. We, therefore, may be able to obtain more useful information regarding the source of the changes by looking at the frequency response of the FBG signal. This technique is broadly discussed in the result and discussion section (\ref{results}).

\begin{figure}
  \centering
    \includegraphics[scale=0.8]{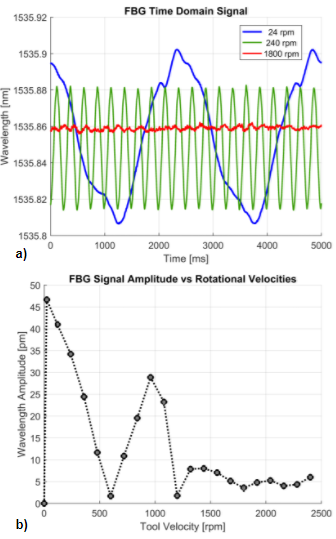}
      \caption{ a) Time domain FBG signal for different tool rotational velocities when the CM is in straight pose, and b) amplitude of the periodic FBG signals for different tool rotational velocities.}
      \label{fig:straight-time}
\end{figure}

\section{Experiments} \label{experiments}

\subsection{Experimental Setup} \label{setup}


The experimental setup (Fig. \ref{fig:setup}) consists of the CM actuation unit with two DC motors (RE16, Maxon Motor Inc.) with spindle drives (GP16, Maxon Motor, Inc.) to pull the actuation cables. Furthermore, the setup has one stepper motor (DMX-UMD-23, Arcus Technology, Inc.) which is used for rotating the debriding tool. A custom C++ interface performs independent velocity or position control of each actuation motor using libraries provided by Maxon Inc. and communicated over a single mini-USB cable.

FBG data is streamed by a dynamic optical sensing interrogator (Micron Optics sm 130) at the maximum frequency of $1$ KHz. Two fibers each with three active areas, associated with two shape sensors embedded into the manipulator, are connected to separate channels of the interrogator.
   
\subsection{Experiments}

As mentioned in section \ref{tools}, optimal debriding performance happens at different ranges of rotational tool velocities in soft and hard lesion removal \cite{alambeigi2016design,alambeigiOsteoNec}. For soft lesion debriding, we reach optimal performance in relatively lower rotational velocities compared to hard lesion debriding. Therefore, we design our experiments to cover a broad range of rotational velocities for the tool (i.e. $0-2400$ rpm). 

In addition, in the real scenario of lesion debriding, first the CM is guided through the screw holes of the acetabular implant by a robotic arm, the debriding tool starts to rotate, and then the cables are actuated to bend the manipulator and make contact with the lesion. Therefore, we need to study the system behavior in two situations: $1)$ the tool is rotating while the manipulator is straight (unbent), and $2)$ the tool is rotating and the manipulator is being bent simultaneously. 

To mimic different stages of the real osteolysis treatment scenario, and to address different debriding tool velocity ranges, we designed two sets of experiments, one when the manipulator is straight (cables are not actuated) and one where the manipulator is continuously bent (actuation cable is pulled by the motor). We repeat both sets of experiments with different tool rotational velocities in the range of $0-2400$ rpm.

\subsubsection{Straight Manipulator} \label{straightExp}
In this set of experiments, we do not actuate the manipulator cables. In other words, no constraints are imposed to the manipulator by the cables. A debriding tool (Fig. \ref{fig:CDM}) is passed through the lumen of the manipulator. The debriding tool is rotated by different rotational velocities in the specified range, and the FBG data of the shape sensors are collected.

\subsubsection{Bending Manipulator}\label{bentExp}
In these experiments, we started to rotate the debriding tool with a constant rotational velocity, and simultaneously bent the manipulator by actuating one of its cables with constant velocity of $0.1$ mm/s (which agrees with optimal actuating velocities suggested in \cite{alambeigi2016design}). The cable is actuated, then released so that the manipulator goes back to its straight pose, and then this bend/unbend procedure is repeated for a second time. The FBG data of the shape sensors are collected in the entire manipulator movement. Various tool rotational velocities in the specified range are experimented.

\section{Results and Discussion} \label{results}

\subsection{Straight Manipulator} \label{straightResult}

Fig \ref{fig:straight-time} shows the result of the experiments when the manipulator is in straight pose (cables are not actuated) and different tool rotational velocities are imposed. Fig. \ref{fig:straight-time}-a shows the FBG time domain response for different tool rotational velocities. It can be observed that the FBG signal is oscillating periodically with a frequency related to the rotational velocity. As the tool rotational velocity increases, the amplitude and frequency of the signal change. More specifically, the frequency of the signal tends to increase when the rotational velocity has increased. Take the FBG signal associated with $240$ rpm  ($4$ rps) tool velocity as an example. The green FBG signal has also oscillated $4$ times in each second, supporting the fact that the rotational tool is harmonically exciting the system with the same frequency as the tool rotates. We can therefore plot the amplitude of the periodic FBG signals against different tool rotational velocities and obtain the natural frequencies of the system \cite{vibration}.

Fig. \ref{fig:straight-time}-b indicates how the amplitude of the FBG signal changes as the tool rotational velocity varies. It can be observed that the amplitude of the signal has two peaks, one at ${\sim}24$ rpm ($\omega_1=2 \pi (0.4)$) and the other ${\sim}960$ rpm ($\omega_2=2 \pi (16)$). These are the natural frequencies of the system, for which large amplitudes of the signal can be observed. It should be noted that the sensor assembly is not rigidly connected to the small channel inside the CM wall (it is only glued at the distal end of the CM \cite{sefati2016fbg}). Therefore, as the tool harmonically excites the system, the sensor assembly and the CM might vibrate differently, since they have different mechanical properties (e.g. mass and moment of inertia). The overall trend of the data in Fig. \ref{fig:straight-time}-a looks very similar to that of a two degrees of freedom system excited by a harmonic force, which has two natural frequencies. In these experiments, we observed that the CM became very unstable with large amplitude oscillations around its natural frequency $\omega_2$ (rotational velocities around $960$ rpm), which can also be confirmed by results of Fig. \ref{fig:straight-time}-b. Therefore, for debriding experiments, rotational velocities far from this value should be chosen to avoid unstable behavior of the CM. On the other side of the rotational velocity spectrum (i.e. low velocities), the CM did not show large amplitudes of oscillation in the experiments, while the amplitude of the FBG signal is large. Therefore, the large FBG data amplitudes for velocities around $24$ rpm may reflect that the shape sensor is oscillating with a high amplitude inside the CM sensor channel (and not the CM itself), indicating that the natural frequency of $\omega_1$ is causing the shape sensor to vibrate with large amplitudes.  

\begin{figure}
  \centering
    \includegraphics[scale=0.78]{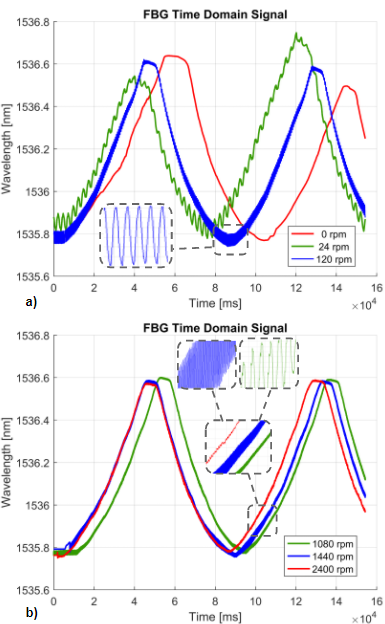}
      \caption{Time domain FBG signal when the CM is being actuated with a) low tool rotational velocities and b) high tool rotational velocities.}
      \label{fig:bent-time}
\end{figure}

\subsection{Bending Manipulator} \label{bentResult}

Fig. \ref{fig:bent-time} demonstrates the time domain FBG signals for different tool rotational velocities while the CM is simultaneously  bending, unbending, and repeating this pattern. Fig. \ref{fig:bent-time}-a and -b each compare the behavior for three low and high rotational velocities, respectively. For lower velocities (\ref{fig:bent-time}-a), we can observe that the FBG wavelength data is repeating a large increase and decrease pattern due to the bending of the CM. In other words, the overall shape of the data is associated with the actual changes in the shape of the CM. In addition, due to the tool rotation, the data is oscillating with a faster rate on top of the changes due to the CM shape (i.e. changes with lower rate). The oscillation frequency gets higher as the tool rotational velocity increases. This behavior indicates the fact that in vibration due to a harmonic force, the system oscillates with the same frequency as the frequency of the harmonic force. 

Another important observation from Fig. \ref{fig:bent-time} is the changes in the amplitude of the oscillations due to tool rotations. This amplitude in generally larger in lower tool velocities (Fig. \ref{fig:bent-time}-a) as compared to higher tool velocities (Fig. \ref{fig:bent-time}-b). We can relate this behavior to the relation between amplitude and tool velocity in Fig. \ref{fig:straight-time}-b, where higher tool velocities in the range of approximately $ 1100-2400$ rpm result in lower oscillation amplitudes in the FBG signal compared to the range of approximately $24-400$ rpm. Moreover, the amplitude of the oscillation varies within a single FBG signal. Take the signal associated with tool rotational velocity $120$ rpm (blue signal) in Fig. \ref{fig:bent-time}-a. The amplitude of oscillation is larger at the beginning where the CM is straight, compared to the peak of the signal where the CM is bent and the FBG wavelength is maximum ( ${\sim} 1536.6$ nm). The oscillation amplitude gets larger again when the CM is being unbent to its straight pose (at time ${\sim}90$ seconds). The larger amplitude when the CM is close to its straight pose is due to the fact that the actuation cables are getting loose at this point and therefore they are not imposing a physical constraint on the CM to limit its oscillations.

\addtolength{\textheight}{-0.7 cm}  
                                  
\subsection{Frequency Domain Analysis}
Although valuable information can be inferred from the time domain FBG signal, when the CM is in dynamic environments (e.g. in the presence of vibration), the frequency domain can potentially provide better information on the behavior of the system. In addition, the frequency domain analysis enables more systematic identification of the system frequency response, which can result in more accurate and exact pre-processing of the data \cite{DSP}. 
Suppose that in addition to shape reconstruction of the CM, we were to obtain other useful information from the FBG signal. As an example, we might be interested to look for any sudden changes in the FBG signal, which might be a sign of colliding with an obstacle in the environment. Simply looking for sudden changes in the FBG data in the presence of vibration will not provide any useful information since the signal is oscillating rapidly due to the harmonic excitation of the tool. This signal is full of rapid changes due to the tool oscillations and therefore, the data needs to be manipulated so that the high frequency changes due to the tool oscillations are filtered and only changes due to the collision stay present in the signal. For this reason, we take a close look at the frequency domain response of the FBG signal. 

The FBG data is a digital signal sampled with sampling frequency ($F_s$) of $1$ KHz, which is denoted as x[n]. The time domain FBG signals for the CM bending experiments presented in Fig. \ref{fig:bent-time} are not necessarily periodic, even though the tool is rotating periodically. This is simply because the shape of the CM is not changing periodically i.e the CM cable is not necessarily actuated periodically. Therefore, we take the Discrete Fourier Transform (DFT) of the time domain signal in order to obtain the frequency content of it. Given N consecutive samples x[n], ($0 \leq n \leq N-1 $) of an aperiodic sequence, the N-point DFT X[k], ($0 \leq k \leq N-1 $) is defined by \cite{DSP}:

\begin{equation}
X[k] = \sum_{n = 0}^{N-1} x[n] e^{-j \frac{2 \pi}{N} kn}
\end{equation}
where X[k] is a function of the discrete frequency index k, which corresponds to a discrete set of frequencies $ \omega_k = (2 \pi /N)$, and $k = 0, 1, ... ,N-1$. The DFT can be efficiently computed using an algorithm called Fast Fourier Transform (FFT) \cite{DSP}. 

\begin{figure}
  \centering
    \includegraphics[scale=0.44]{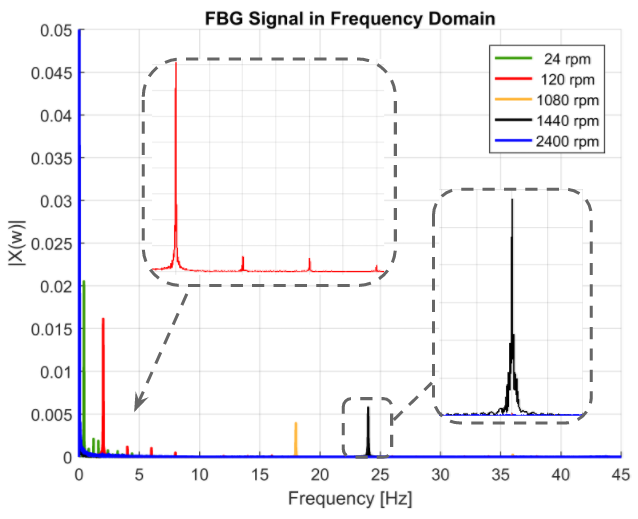}
      \caption{FBG signal in frequency domain by taking Fast Fourier Transform (FFT) of the time domain signal for different rotational velocities.}
      \label{fig:bent-freq}
\end{figure}

We take the \textit{fft} function in MATLAB on the FBG data recorded from the bending CM experiments (section \ref{bentExp}). According to the Sampling Theorem \cite{DSP}, and given that the FBG signal is sampled with sampling frequency of $1$ KHz, we can recover the frequency content of the signal up to $500$ Hz. The FFT algorithm, therefore, will produce the frequency content of the signal up to $500$ Hz. However, for the tool rotational velocities of interest, the maximum frequency content of the FBG signal will not exceed ${\sim}40$ Hz. As a result, only the frequency content of the signal up to this value will be studied. Moreover, the FFT algorithm produces complex numbers as output, the magnitude of which is plotted against frequency in Fig. \ref{fig:bent-freq}. 

We can observe from Fig. \ref{fig:bent-freq} that the amplitude of $X(\omega)$ has larger values where a specific frequency is dominant in a signal. All the plotted signals indicate a base frequency ($F_b$) at ${\sim}0.01$ Hz, which is the frequency due to actual changes in the CM shape. In addition, for each signal we observe a dominant higher frequency with large amplitude which is associated with the oscillations due to the tool vibrations. Take the signal associated with the $120$ rpm tool velocity as example (red plot). The FFT results is showing a dominant peak amplitude at ${\sim}2$ Hz frequency (called the \textit{fundamental frequency}), as well as some small peaks at positive integer multiples of this frequency (called the \textit{harmonics} of the fundamental frequency). For the purpose of our analysis, it suffices to look at the fundamental frequencies of the signal. One other important note from Fig. \ref{fig:bent-freq} is that no peak is observed for the highest tool velocity ($2400$ rpm) other than $F_b$. The FBG time domain signal for this tool velocity in Fig. \ref{fig:bent-time} indicates very small amplitudes of oscillation, which is why the FFT does not capture any peaks for this oscillation frequency. For the sake of the debriding, this means that the FBG data is not affected much by harmonic forces at high tool rotational velocities (${\sim}2400$ rpm).

\section{CONCLUSION} \label{conclusion}
We investigated the behavior of FBG-based shape sensors inside a CM in dynamic situations, where a harmonic force (rotating tool) is exciting the system. Experimental results indicated the feasibility of using FBGs in such environments i.e. high frequency oscillations due to harmonic forces can be captured by FBGs. In addition, time domain analysis of the FBG data identified the two natural frequencies of the system, each of which were associated with large oscillations of either the CM or the shape sensor. This behavior can be justified by the design of the system i.e. the shape sensor is not entirely connected to the CM, rather it is only glued at the distal end of the CM, therefore it can vibrate differently compared to the CM. The time domain FBG signal is associated with the changes in the shape of the CM, vibration of the shape sensor, possible interactions with external objects, as well as noise. Therefore, a more intelligent pre-processing (filtering) of the data can avoid possible loss of valuable information from the FBG data. For this reason, we analyzed the frequency domain of the FBG signal, which gave more intuition on the frequency contents of the signal in presence of vibration. Based on this information, we can design a filter to specifically filter out the effect of harmonic forces. This can be achieved via a band-stop filter to cancel out the frequency contents of the signal due to vibration. Future work will focus on design of such filters, as well as study of the interactions with external objects, while the sensor is harmonically excited by tools.






\end{document}